\documentclass[sigconf]{acmart}

\usepackage[final]{showlabels}
\usepackage{booktabs} 
\usepackage{amsfonts}
\usepackage{amsmath}
\usepackage{graphicx}
\usepackage{paralist}
\usepackage{balance}
\usepackage[T1]{fontenc}
\usepackage{subfigure}
\usepackage{algorithm2e}
\usepackage{amsthm}
\usepackage{hyperref}
\usepackage{xcolor}
\usepackage{pifont}
\usepackage{multirow}

\usepackage{mathtools}
\usepackage{listings} 
\usepackage{bbm}
\usepackage{adjustbox}
\usepackage{array}
\usepackage{enumitem}

\usepackage{bbding}

\usepackage[skip=0.1\baselineskip]{caption}
\hypersetup{
    colorlinks=false,
    linkcolor={red!20!black},
    citecolor={green!20!black},
    urlcolor={blue!20!black}
}

\copyrightyear{2021}
\acmYear{2021}
\setcopyright{acmcopyright}
\acmConference[CIKM '21]{Proceedings of the 30th ACM International Conference on Information and Knowledge Management}{November 1--5, 2021}{Virtual Event, QLD, Australia}
\acmBooktitle{Proceedings of the 30th ACM International Conference on Information and Knowledge Management (CIKM '21), November 1--5, 2021, Virtual Event, QLD, Australia}
\acmPrice{15.00}
\acmDOI{10.1145/3459637.3482475}
\acmISBN{978-1-4503-8446-9/21/11}

\newcommand{\spara}[1]{\smallskip\noindent{\bf #1}}

\newcommand{\rurl}[1]{\href{http://#1}{#1}}
\newcommand{\cmark}{\ding{51}}
\newcommand{\xmark}{\ding{55}}
\newcommand{\rot}[1]{\rlap{\rotatebox{29}{#1}~}}
\newcommand{\tablevert}[1]{\rlap{\rotatebox{90}{#1}~}}

\title{SciClops: Detecting and Contextualizing Scientific Claims \mbox{for Assisting Manual Fact-Checking}}

\author{Panayiotis Smeros}
\affiliation{%
  \institution{EPFL}
  \city{Lausanne}
  \country{Switzerland}
}
\email{panayiotis.smeros@epfl.ch}

\author{Carlos Castillo}
\affiliation{%
	\institution{Universitat Pompeu Fabra}
	\city{Barcelona}
	\country{Catalunya, Spain}
}
\email{chato@acm.org}

\author{Karl Aberer}
\affiliation{%
	\institution{EPFL}
	\city{Lausanne}
	\country{Switzerland}
}
\email{karl.aberer@epfl.ch}

\renewcommand{\shortauthors}{P. Smeros et al.}

\begin{document}

\fancyhead{}
\sloppy

\begin{abstract}
This paper describes SciClops, a method to help combat online scientific misinformation.
Although automated fact-checking methods have gained significant attention recently, they require pre-existing ground-truth evidence, which, in the scientific context, is sparse and scattered across a constantly-evolving scientific literature.
Existing methods do not exploit this literature, which can effectively contextualize and combat science-related fallacies. 
Furthermore, these methods rarely require human intervention, which is essential for the convoluted and critical domain of scientific misinformation.

SciClops involves three main steps to process scientific claims found in online news articles and social media postings: extraction, clustering, and contextualization.
First, the extraction of scientific claims takes place using a domain-specific, fine-tuned transformer model.
Second, similar claims extracted from heterogeneous sources are clustered together with related scientific literature using a method that exploits their content and the connections among them.
Third, check-worthy claims, broadcasted by popular yet unreliable sources, are highlighted together with an enhanced fact-checking context that includes related verified claims, news articles, and scientific papers.
Extensive experiments show that SciClops tackles sufficiently these three steps, and effectively assists non-expert fact-checkers in the verification of complex scientific claims, outperforming commercial fact-checking systems.
\end{abstract}

%


\maketitle


\section{Introduction}\label{sec:introduction}

\begin{figure}[t]
	\centering
	\includegraphics[width=.91\columnwidth]{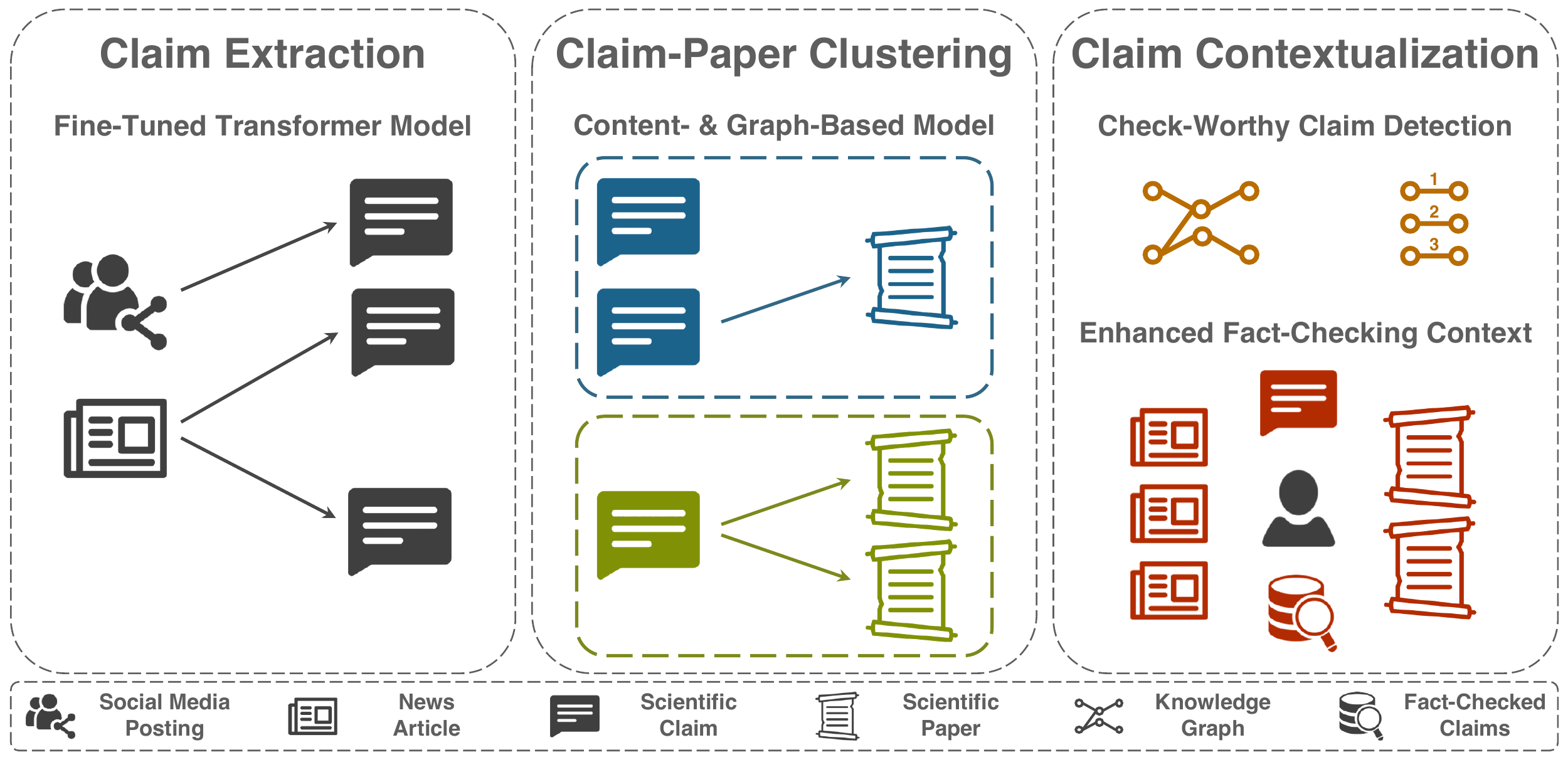}
	\caption{Overview of SciClops including the three methods for extraction (\S\ref{sec:extraction}), clustering (\S\ref{sec:clustering}), and contextualization (\S\ref{sec:verification}) of scientific claims.}
	\label{fig:SciClops_Overview}
\end{figure}

\begin{table*}[t]
	\setlength{\tabcolsep}{5pt}
	\scriptsize
	\centering
	\caption{Approaches for Extraction, Clustering, and Contextualization as proposed by selected references}
	\label{table:ce_techniques}
	\begin{tabular}{{l}l*{28}c}
		\toprule
		&
		\rot{Fact-Checking Portals} &
		\rot{\citet{DBLP:conf/kdd/HassanALT17}} &
		\rot{\citet{DBLP:conf/www/PopatMSW17}} &
		\rot{\citet{DBLP:conf/naacl/JaradatGBMN18}} &
		\rot{\citet{DBLP:conf/acl/ShaarBMN20}} &
		\rot{\citet{DBLP:conf/www/HansenHASL19}} &

		\rot{\citet{DBLP:conf/emnlp/ZlatkovaNK19}} &
		\rot{\citet{DBLP:journals/pvldb/Karagiannis0PT20}} &
		\rot{\citet{DBLP:conf/jcdl/PintoWB19}} &
		
		\rot{\citet{DBLP:conf/icwsm/PavlloP018}} &
		\rot{\citet{DBLP:conf/www/Smeros0A19}} &
		\rot{\citet{DBLP:conf/coling/LevyBHAS14}} &
		\rot{\citet{DBLP:conf/emnlp/StabMSRG18}} &
		\rot{\citet{DBLP:conf/cikm/PatwariGB17}} &
		\rot{\citet{DBLP:conf/ijcai/LippiT15}} &
		\rot{\citet{DBLP:conf/www/JiangBI020}} &		
		\rot{\citet{DBLP:conf/acl/ReimersSBDSG19}} &
		
		\rot{\citet{DBLP:conf/aaai/YaoM019}} &
		\rot{\citet{DBLP:journals/pvldb/ZhouCY09}} &
		\rot{\citet{DBLP:conf/nips/HamiltonYL17}} &
		\rot{\citet{DBLP:conf/aaai/WangCWP0Y17}} &
		\rot{\citet{duongchithang}} &
		
		\rot{\citet{DBLP:conf/coling/KochkinaLZ18}} &
		\rot{\citet{DBLP:conf/www/ShaoCFM16}} &
		\rot{\citet{doi:10.1371/journal.pone.0128193}} &
		\rot{\citet{DBLP:conf/naacl/NadeemFXMG19}} &
		\rot{\citet{DBLP:conf/wsdm/Gad-Elrab0UW19}} &
		\rot{\citet{DBLP:conf/iclr/ChenWCZWLZW20}} &
		
		\textbf{SciClops}\\
		\midrule
		
		\textbf{Extraction}\\
		Weak Supervision
		&\cmark&\xmark&\xmark&\xmark&\xmark&\cmark&\xmark&\xmark&\xmark&\cmark&\cmark&\xmark&\xmark&\xmark&\xmark&\xmark&\xmark&-          &-          &-          &-          &-          &-         &-          &-          &-          &-          &-          &\cmark \\
		Traditional ML Model		  
		&\xmark&\cmark&\cmark&\xmark&\xmark&\xmark&\cmark&\cmark&\cmark&\xmark&\xmark&\cmark&\cmark&\cmark&\cmark&\xmark&\xmark&-          &-          &-          &-          &-         &-          &-          &-          &-          &-          &-          &\cmark \\
		Neural ML Model
		&\xmark&\xmark&\xmark&\cmark&\cmark&\cmark&\xmark&\xmark&\xmark&\xmark&\xmark&\xmark&\xmark&\xmark&\xmark&\cmark&\cmark&-          &-          &-          &-          &-          &-         &-          &-          &-          &-          &-          &\cmark \\
		\textbf{Clustering}\\
		Text Modality
		&\cmark&\cmark&\xmark&\xmark&\xmark&\xmark&\xmark&\xmark&\cmark&-          &-          &-          &-          &-          &-          &-         &\cmark&\cmark&\cmark&\cmark&\xmark&\xmark&-          &-          &-         &-          &-          &-          &\cmark \\
		Graph Modality
		&\xmark&\xmark&\xmark&\xmark&\xmark&\xmark&\cmark&\xmark&\cmark&-          &-          &-          &-          &-         &-          &-          &\xmark&\cmark&\cmark&\cmark&\cmark&\cmark&-          &-          &-         &-          &-          &-          &\cmark \\
		Bipartite Clusters
		&\xmark&\xmark&\xmark&\xmark&\xmark&\xmark&\xmark&\xmark&\xmark&-          &-          &-          &-          &-         &-          &-         &\xmark&\xmark&\xmark&\xmark&\xmark&\cmark&-           &-          &-         &-          &-          &-          &\cmark \\
		\textbf{Contextualization}\\
		Ground-Truth KBs
		&\cmark&\cmark&\cmark&\xmark&\cmark&\xmark&\cmark&\cmark&\xmark&-          &-          &-          &-          &-          &-          &-         &-          &-          &-          &-          &-          &-         &\cmark&\cmark&\cmark&\cmark&\cmark&\cmark&\cmark \\
		Priority Ranking
		&\xmark&\xmark&\xmark&\cmark&\cmark&\cmark&\xmark&\cmark&\xmark&-          &-          &-          &-          &-          &-          &-         &-          &-          &-          &-          &-          &-         &\xmark&\xmark&\xmark&\cmark&\xmark&\xmark&\cmark \\
		Scientific Context
		&\cmark&\xmark&\xmark&\xmark&\xmark&\xmark&\xmark&\xmark&\cmark&-          &-          &-          &-          &-          &-          &-         &-          &-          &-          &-          &-          &-         &\xmark&\xmark&\xmark&\xmark&\xmark&\xmark&\cmark \\
		\bottomrule
	\end{tabular}
\end{table*}

Although the amount of news at our disposal seems to be ever-expanding, traditional media companies and professional journalists remain the key to the production and communication of news.
The way in which news is disseminated has become more intricate than in the past, with social media playing a fundamental role~\cite{pewresearch2018}.

The ephemeral, fast-paced nature of social media, the brevity of the messages circulating on them, the short attention span of their users, their preference for multimedia rather than textual content, and in general the fierce competition for attention, has forced journalists to adapt in order to survive in the attention economy~\cite{doi:10.1080/16522354.2018.1527521}.
As a consequence, news outlets are increasingly using catchy headlines, as well as outlandish and out-of-context claims that perform well in terms of attracting eyeballs and clicks~\cite{DBLP:conf/asunam/RonyHY17}.

When mainstream news media communicate scientific content to the public, the situation is by no means different~\cite{Scheufele14040}.
Oversimplified scientific claims are rapidly shared in social media, while the scientific evidence that may support or refute them remains absent or locked behind pay-walled journals.
For instance, on March 11th, 2020, an article in \emph{The Lancet Respiratory Medicine} theorized that nonsteroidal anti-inflammatory drugs such as Ibuprofen could worsen COVID-19 symptoms~\cite{fang2020patients}.
Without referencing explicitly to this article, but motivated by it, the Minister of Health of France posted on Twitter, advising people to avoid Ibuprofen when possible.\footnote{\url{https://twitter.com/olivierveran/status/1238776545398923264}
}
His message was re-posted nearly $43K$ times and liked nearly $40K$ times.
In contrast, a World Health Organization's message posted four days later, which insisted Ibuprofen was safe, was re-posted only $7.5K$  times and liked only $8.5K$ times.\footnote{\url{https://twitter.com/WHO/status/1240409217997189128}
}

Fact-checking portals such as \rurl{ScienceFeedback.co}, among others, work closely with domain experts and scientists to debunk misinformation and bring nuance to potentially misleading claims.
This remains, however, a labor-intensive and time-consuming task~\cite{hassan2015quest}.
On the other hand, despite misinformation circulating online exceeding the capacity of manual fact-checking, traditional news outlets are skeptical towards adopting fully-automated methods \cite{deutschewelle2020}.
Their main concern is that such tools provide poorly-interpretable evidence (according to the journalistic standards), and any false judgment can lead to a downfall of the outlet's reputation.
Indeed, even big tech companies were forced to suspend automated fact-checking features due to similar criticism from news outlets \cite{poynter2018}.
Hence, the consensus regarding the usage of automation in journalism is that it should assist but not replace journalists and news consumers when they validate the veracity of news, enabling the movement onward the era of citizen journalism \cite{DBLP:journals/corr/abs-2103-07769}.

%

Our work focuses on scientific claims in news articles and social media postings.
As scientific claims, we consider \emph{sentence-level segments that involve one or more scientific entities and are eligible for fact-checking}.
For example, the sentence \emph{``Ibuprofen can worsen COVID-19 symptoms''} is a scientific claim because it involves two scientific entities (\emph{Ibuprofen} and \emph{COVID-19}) and implies a causal relation between them.
To increase the coverage of our definition, we bound neither the number of entities nor the type of relation between them.
%
%
%
Such non-deterministic definition 
makes the detection of scientific claims a challenging task, even for human annotators (details in \S\ref{sec:results_extraction}).
To address this task and enable the discovery of complex-structured claims, there is a need for advanced language models which are fine-tuned with domain-specific knowledge. 
%

%
%

%
%
%

%

Once we identify candidate scientific claims, we seek evidence that proves or contradicts them via \emph{contextualization}, i.e., via building an enhanced context of trustworthy information.
In the scientific domain, the appropriate context consists of related scientific papers.
Grouping similar claims and linking them to related scientific literature is a complex task, to a large extent because of the different nature of the items that we are seeking to connect (i.e., social media postings, news articles, and scientific papers).
These contain key passages that determine such connections, but are fundamentally different in terms of:
\begin{inparaenum}[i)]
	\item verbosity, ranging from character-limited postings to extended scientific papers, and
	\item complexity, ranging from a ``social media friendly'' style of writing to the more formal registry of journalism and academic writing.
\end{inparaenum}

Finally, since there is a plethora of controversial claims (especially in the times of a pandemic), there is a need for a check-worthiness ranking that considers the prevalence and the reliability of the broadcasting medium.
Providing a scientific context enables non-expert fact-checkers to verify claims with more precision than commercial fact-checking systems, and more confidence since the provided context is fully-interpretable (details in \S\ref{sec:results_verification}).

%

\spara{Our Contribution.}
In this paper we describe \emph{SciClops} (Figure~\ref{fig:SciClops_Overview}), a method
%
%
to assist manual verification of dubious claims, in scientific fields with open-access literature and limited fact-checking coverage.
The technical contributions we introduce are the following:
\begin{itemize}[leftmargin=*]
	\item pretrained and fine-tuned transformer-based models for scientific claim extraction from news and social media ($\S$\ref{sec:extraction});
	\item multimodal, joint clustering models for claims and papers that utilize both content and graph information ($\S$\ref{sec:clustering});
	\item methods for ranking check-worthy claims using a custom knowledge graph, and methods for creating enhanced scientific contexts to assist manual fact-checking ($\S$\ref{sec:verification}); and
	\item extensive experiments involving expert and non-expert users, strong baselines and commercial fact-checking systems ($\S$\ref{sec:results}).
\end{itemize}

\section{Related Work}\label{sec:relatedwork}

\emph{Fact Checking Portals} in general (\rurl{Snopes.com}), political (\rurl{PolitiFact.com}), and scientific (\rurl{ScienceFeedback.co}) domains employ specialized journalists who manually: 
\begin{inparaenum}[i)]
	\item detect suspicious claims (extraction),
	\item discover variants of these claims published in social and news media (clustering), and
	\item find the appropriate prism under which they assess their credibility (contextualization).
\end{inparaenum}
We summarize some automated methods tackling these steps in Table~\ref{table:ce_techniques}.

\spara{Claim Extraction.}
%
%
%
%
%
On \emph{weakly supervised models}, \citet{DBLP:conf/icwsm/PavlloP018} and \citet{DBLP:conf/www/Smeros0A19} generate complex rule-based heuristics to extract quotes from, respectively, general and scientific news articles.

On \emph{traditional ML models}, \citet{DBLP:conf/coling/LevyBHAS14} and \citet{DBLP:conf/emnlp/StabMSRG18} propose learning models for claim detection and argument mining and introduce publicly available datasets, which we utilize to train our extraction models (details in \S\ref{sec:results_extraction}).
\citet{DBLP:conf/kdd/HassanALT17} and \citet{DBLP:conf/www/PopatMSW17} propose claim classification models that use the aforementioned fact-checking portals to verify political claims, while \citet{DBLP:conf/cikm/PatwariGB17} and \citet{DBLP:conf/ijcai/LippiT15} propose, respectively, an ensemble and a context-independent model for claim extraction.
Finally, \citet{DBLP:conf/emnlp/ZlatkovaNK19} propose a claim extraction model for images, \citet{DBLP:journals/pvldb/Karagiannis0PT20} propose a framework for statistical claims verification, and \citet{DBLP:conf/jcdl/PintoWB19} propose a method for identifying pairwise relationships between scientific entities.

On \emph{neural ML models}, \citet{DBLP:conf/naacl/JaradatGBMN18} and \citet{DBLP:conf/acl/ShaarBMN20} detect and rank previously fact-checked claims using deep neural models, while \citet{DBLP:conf/www/HansenHASL19} also train a neural ranking model for check-worthy claims using weak supervision.
Furthermore, \citet{DBLP:conf/www/JiangBI020} use contextualized embeddings to factor fact-checked claims, while \citet{DBLP:conf/acl/ReimersSBDSG19} use also contextualized embeddings for claim extraction and clustering.
Finally, \texttt{CheckThat!} Lab \cite{DBLP:conf/clef/Barron-CedenoEN20} features claim extraction and check-worthiness tasks which are oriented towards political debates in social media platforms.

While the other approaches cover the cases of political, statistical, and visual claims, our approach provides the first dedicated solution for scientific claims. 
Given the complex nature of the scientific claims in terms of structure and vocabulary, our approach is based on advanced language models with contextualized embeddings that are fine-tuned with domain-specific knowledge.
Furthermore, our approach works with arbitrary input text, e.g., from social media postings, blog posts, or news articles.

%
%
%

%

%


\spara{Claim-Paper Clustering.}
Since our data contains multimodal information (the textual representation of claims and papers and the interconnections between them), we present multimodal clustering approaches that combine text and graph data modalities.
%

%
%
\citet{DBLP:conf/aaai/YaoM019} propose a unified convolutional network of terms and documents, 
while
\citet{DBLP:journals/pvldb/ZhouCY09} use weighted graphs that encode the attribute similarity of the clustered nodes.
%
%
%
\citet{DBLP:conf/nips/HamiltonYL17} introduce a methodology for jointly training embeddings based on text and graph information, while
\citet{DBLP:conf/acl/ReimersSBDSG19} apply a numerical clustering on top of such embeddings.
Finally, \citet{DBLP:conf/aaai/WangCWP0Y17} propose a technique for training network embeddings that preserves the communities (clusters) of a graph, while \citet{duongchithang} provide interpretable such embeddings.

%

In our approach, we jointly cluster scientific claims and referenced papers, using both content and graph information.
To the best of our knowledge, this is the first approach that deals with heterogeneous passages in terms of length and vocabulary type, which are also interconnected through a bipartite graph. 

\spara{Claim Contextualization.}
%
%
In addition to the extraction methods described above, the majority of which also provide contextualization/verification techniques (details in Table~\ref{table:ce_techniques}),
\citet{DBLP:conf/coling/KochkinaLZ18} and \citet{DBLP:conf/www/ShaoCFM16} propose methods for automatic rumor verification using well-known fact-checking portals.
\citet{doi:10.1371/journal.pone.0128193}, \citet{DBLP:conf/naacl/NadeemFXMG19}, and \citet{DBLP:conf/iclr/ChenWCZWLZW20} use Wikipedia for fact-validation, while
\citet{DBLP:conf/wsdm/Gad-Elrab0UW19} use custom knowledge graphs for generating interpretable explanations for candidate facts.
%

%
While other approaches describe this step as \emph{``verification''}, since essentially they lookup a claim in a ground-truth knowledge base, we consider the general case in which claims rarely appear in such knowledge bases.
As we observe in \S \ref{sec:results_verification}, this is a pragmatic assumption since the majority of the fact-checking effort targets non-scientific topics.
%
As the verification of scientific claims is typically more demanding than other types of claims (e.g., \rurl{ScienceFeedback.co} has built an entire peer-reviewing system for this purpose), we propose a methodology that contextualizes claims based on related scientific literature and ranks them based on the prevalence and the reliability of the broadcasting medium.

%

%


\section{Claim Extraction}\label{sec:extraction}

%
We address claim extraction as a classification problem at the sentence level, i.e., we want to distinguish between claim-containing and non-containing sentences.
Below, we present the baseline and the advanced extractors that we evaluate in \S\ref{sec:results_extraction}.
%
%

\subsection{Baseline Extractors}\label{sec:heuristic-extractors}
We implement several baseline extractors that cover most of the related work on claim extraction described in \S\ref{sec:relatedwork}:
\begin{inparaenum}[i)]
	\item two complex heuristics which are used by state-of-the-art \emph{weakly supervised models} \cite{DBLP:conf/icwsm/PavlloP018, DBLP:conf/www/Smeros0A19};
	\item an off-the-shelf classifier trained with standard textual features which is used by state-of-the-art \emph{traditional ML models} \cite{DBLP:conf/kdd/HassanALT17, DBLP:conf/ijcai/LippiT15}; and
	\item a transformer model which is used by state-of-the-art \emph{neural ML models} \cite{DBLP:conf/acl/ReimersSBDSG19, DBLP:conf/acl/ShaarBMN20}.
\end{inparaenum}

%
%

\subsubsection{Grammar-Based Heuristic}

The usage of reporting verbs such as ``say,'' ``claim,'' or ``report,'' is a typical element of pattern-matching heuristics for finding claims.
Another element is the usage of domain-specific vocabulary; in the scientific context, common verbs in claims include ``prove'' and ``analyze.''
Thus, we compile a seed set of such verbs, which we extend with synonyms from WordNet~\cite{DBLP:journals/cacm/Miller95}.
In the following, we refer to this set of reporting verbs as $RV$.

Scientific claims fundamentally refer to scientific studies, scientists or, more generally, scientific notions.
Thus, we employ a shortlist of nouns related to studies and scientists (including ``survey'' or ``researcher'').
In the following, we refer to this set of nouns, together with the set of Person and Organization entities, as $E$.

Finally, to capture the syntactic structure of claims, we obtain part-of-speech tags from the candidate claim-containing sentences.
Using this information, we construct a series of complex expressions over \emph{classes of words} such as the following:
\begin{center}
\small
$(\mathit{root}(s) \in \mathit{RV}) \land ((\mathit{nsubj}(s) \in E)  \lor (\mathit{dobj}(s) \in E)) \implies (s \in \mathit{Claims})$
\end{center}
where $s$ is a sentence, \emph{root}$(.)$ returns the root verb of the syntactic tree of a sentence, \emph{nsubj}$(.)$ returns the nominal subject, and \emph{dobj}$(.)$ the direct object of a sentence.

\subsubsection{Context-Based Heuristic}
This heuristic is based on a frequent non-syntactic pattern, which is quite evident in our data: if an article is posted on social media, then its central claim is typically re-stated or minimally paraphrased in the postings.
%
%
%
%
We investigate pairs (s,p) of candidate sentences $s$, extracted from news articles, and postings $p$, referencing these news articles.
Our heuristic has the form:
\begin{center}
	$(\exists p: \emph{sim}(s,p) \cdot \emph{pop}(p) \ge \emph{threshold}) \implies (s \in \emph{Claims})$
\end{center}

where $\emph{sim}(s,p)$ denotes the \emph{cosine similarity} between the embeddings representations of  $s$ and $p$, and $\emph{pop}(p)$ denotes the normalized popularity of $p$, i.e., the raw popularity of $p$ over  the sum of the popularity of all the $p$'s that refer to $s$.
As popularity, we consider the sum of the \emph{re-postings} and \emph{likes}.
Finally, \emph{threshold} is a hyper-parameter of our heuristic, which in our implementation is fixed to $0.9$, yielding a good compromise of precision and recall.
We note that this is the only proposed extractor that is not purely content-based since it also requires contextual information.

%
%
%
%
%


%
%

\subsubsection{Random Forest Classifier}
%
To train this classifier, we apply a standard text-preprocessing pipeline, including stop-words removal and part-of-speech tagging. 
Then, we transform the candidate claim-containing sentences into embeddings by averaging the word embeddings provided by GloVe \cite{DBLP:conf/emnlp/PenningtonSM14}.
As we see in our evaluation (\S\ref{sec:results_extraction}), this classifier performs better than the aforementioned baselines; we also note that, compared to the complex transformer models, it is substantially less intensive in terms of computational resources and training time needed.

%
%
%



\subsubsection{BERT Model}
One of the most successful state-of-the-art approaches to several NLP tasks, including classification, is the \emph{transformer model}~\cite{DBLP:conf/naacl/DevlinCLT19}.
In our implementation we use the well-known model \emph{BERT} and particularly its version named \emph{bert-base-uncased} \cite{DBLP:journals/corr/abs-1910-03771}.
The configuration parameters of the model are those suggested in a widely used software release of this model.%
\footnote{\url{https://huggingface.co/bert-base-uncased}}
%
%

As the last layer of the transformer architecture of \emph{BERT} (and the variants we introduce next), we add a standard binary classification layer with two output neurons, which we train using the datasets described in \S\ref{sec:results_extraction}.
During the training, we keep the rest of the layers of the model frozen at their initial parameters.

\subsection{Fine-Tuned Transformer Extractors}\label{sec:sentence_classifier}

Since \emph{BERT} is originally trained on the generic corpus of Wikipedia, the word representations it generates are also generic.
However, scientific claim extraction is a downstream task, where the model has to recognize patterns of a more narrow domain.
Thus, we introduce three variants of \emph{BERT} with domain-specific fine-tuning namely, \emph{SciBERT}, \emph{NewsBERT} and \emph{SciNewsBERT}:
\begin{itemize}[leftmargin=*]
	\item \emph{SciBERT} is pretrained on top of \emph{BERT} with a corpus from \rurl{SemanticScholar.org} containing \textasciitilde$1M$ papers \cite{DBLP:conf/emnlp/BeltagyLC19}. \emph{SciBERT} has its own vocabulary that is built to best match the scientific domain.
	\item \emph{NewsBERT} is a new model that we introduce, built on top of \emph{BERT} and pretrained on a freely-available corpus of \textasciitilde$1M$ headlines published by the Australian Broadcasting Corporation \cite{DVN/SYBGZL_2018}.
	\item \emph{SciNewsBERT} is also a new model that is pretrained like \emph{NewsBERT}, albeit, it is built on top of \emph{SciBERT} instead of \emph{BERT}.
\end{itemize}
For training \emph{NewsBERT} and \emph{SciNewsBERT} we employ the standard tasks for training \emph{BERT}-like models:
\begin{inparaenum}[i)]
	\item \emph{Masked Language Modeling}, where the model has to predict the randomly masked words in a sequence of text, and
	\item \emph{Next Word Prediction}, where the model has to predict the next word, given a set of preceding words.
The hyper-parameters used for training the models are the default proposed by the software release referenced above. 
\end{inparaenum}
Since both \emph{NewsBERT} and \emph{SciNewsBERT} need substantial computational power and training time, we make them publicly available for research purposes (\S\ref{sec:conclusions}).

\section{Claim-Paper Clustering}\label{sec:clustering}

Contextualizing scientific claims requires to connect them with related scientific papers. 
To achieve this, our approach employs a clustering methodology.
%
The clusters, composed of a mixture of claims and papers, must have high semantic coherence and ideally maintain the connections that exist between some of these claims and papers.
These implicit connections are hyperlinks starting from news articles and social media postings containing these claims and ending on referenced papers, forming a sparse bipartite graph.

%
The clustering methods that we employ are:
\begin{inparaenum}[i)]
	\item \emph{Content-Based} methods on top of either the raw text or an embeddings representation of the passages,
	\item \emph{Graph-Based} methods on top of the bipartite graph between the claims and the papers, or
	\item \emph{Hybrid} methods that combine the \emph{Content-Based} and the \emph{Graph-Based} methods.
\end{inparaenum}
Furthermore, we consider both soft (overlapping) clustering (i.e., passages can belong to more than one cluster), and hard (non-overlapping) clustering (i.e., passages must belong to exactly one cluster).
The notation used in this section is summarized in Table~\ref{table:clustering_notation}.
%

\subsection{Content-Based Clustering}\label{sec:content-based_clustering}
Our baseline is content-based (topic) clustering.
According to this approach, we assume that claims and papers are represented in the same latent space,
in which we compute topical joint clusters.
This approach does not consider the interconnections (i.e., the bipartite graph) between the claims and the papers.

For topic modeling, we use \emph{Latent Dirichlet Allocation} (\emph{LDA}), an unsupervised statistical model that computes a soft topic clustering of a given set of passages \cite{DBLP:journals/jmlr/BleiNJ03}.
We also use \emph{Gibbs Sampling Dirichlet Mixture Model} (\emph{GSDMM}), which assumes a hard topic clustering and is more appropriate for small passages such as claims \cite{DBLP:conf/sigir/LiWZSM16}.
When the passages are projected in an embeddings space, we use either the generic \emph{Gaussian Mixture Model} (\emph{GMM}), which computes a soft clustering by combining multivariate Gaussian distributions \cite{DBLP:reference/bio/Reynolds09}, or 
\emph{K-Means} \cite{DBLP:journals/tit/Lloyd82}, which computes a hard clustering.
Finally, we test these methods with and without reducing the embeddings dimensions using
\emph{Principal Component Analysis} (\emph{PCA}) \cite{doi:10.1080/14786440109462720}.

\begin{table}[t]
	\footnotesize
	\centering
	\caption{
		Clustering notation. The embeddings dimension (\emph{dim}) of our models is \emph{300}.
		Matrix \emph{L} has a \emph{1} in position (\emph{c}, \emph{p}), iff a news article or a social media posting containing claim \emph{c} has a hyperlink to paper \emph{p}.
		Each row of the clustering matrices ($\emph{C}^{\prime}$ and $\emph{P}^{\prime}$) contains the probability of a claim or a paper to belong to a cluster; for hard clustering it is ``one-hot'', i.e., it has a single non-zero element, and for soft clustering it is a general probability distribution.
	}
	\label{table:clustering_notation}
	\begin{tabular}{ll}
		\toprule
	 	Symbol & Description \\
		\midrule
		$\emph{C} \in \mathbb{R}^{|\operatorname{claims}| \: \times \: \operatorname{dim}}$ & initial claims matrix \\
		$\emph{P} \in \mathbb{R}^{|\operatorname{papers}| \: \times \: \operatorname{dim}}$ & initial papers matrix \\
		$\emph{L} \in {\{0, 1\}}^{|\operatorname{claims}| \: \times \: |\operatorname{papers}|}$  & interconnection matrix \\
		$\emph{C}^{\prime} \in {[0, 1]}^{|\operatorname{claims}| \: \times \: |\operatorname{clusters}|}$  & final claims clustering matrix \\
		$\emph{P}^{\prime} \in {[0, 1]}^{|\operatorname{papers}| \: \times \: |\operatorname{clusters}|}$  & final papers clustering matrix \\
		$\emph{f}_{\emph{C}} \colon \emph{C} \to \emph{C}^{\prime}$ & non-linear neural transformation \\
		$\emph{f}_{\emph{P}} \colon \emph{P} \to \emph{P}^{\prime}$ & non-linear neural transformation \\
		$\left\Vert.\right\Vert_{F}$ & \emph{Frobenius Norm} \\
		\bottomrule
	\end{tabular}
\end{table}

\subsection{Graph-Based Clustering}\label{sec:graph-based_clustering}
Since our data is multimodal, an alternative to pure \emph{Content-Based} clustering is pure \emph{Graph-Based} clustering.
%
%
%
%
%
%
%
We define this problem as an optimization problem, introducing an appropriate loss function that we want to minimize.
Our goal is to compute the optimal clusters $\emph{C}^{\prime}$ and $\emph{P}^{\prime}$, and our evaluation criterion is the extent to which $\emph{C}^{\prime}$ and $\emph{P}^{\prime}$ fit with the interconnection matrix $\emph{L}$.
Hence, we propose the following loss function:
\begin{center}
$\emph{loss} = \left\Vert\emph{C}^{\prime} - \emph{L}\emph{P}^{\prime}\right\Vert_{F} \label{eq:1}$
\end{center}
This loss function is also known as the \emph{Reconstruction Error} and is commonly used in \emph{Linear Algebra} for factorization and approximation problems. 
By applying this loss function, we force $\emph{C}^{\prime}$ and $\emph{P}^{\prime}$ to be aligned with $\emph{L}$: the claims that appear in a news article should belong to the same cluster as the papers referenced by this article.

A degenerate solution to the problem, if we use only this loss function, is a uniform clustering for both claims and papers.
The loss is minimized, but the clustering is useless, because the probability of any claim and any paper to belong to any cluster is uniform.
%
To overcome this problem, we exploit the following technique that is widely used in image processing \cite{DBLP:journals/tip/LefkimmiatisBU12}.

In row-stochastic matrices (i.e., matrices that each row sums to $1$), a uniform soft clustering has lower \emph{Frobenius Norm} than a non-uniform clustering.
Consequently, any hard clustering has the maximum possible \emph{Frobenius Norm}.
Thus, we introduce a regularizer that imposes non-uniformity on the clusters by penalizing low \emph{Frobenius Norms} for $\emph{C}^{\prime}$ and $\emph{P}^{\prime}$:
\begin{center}
$\emph{regularizer} =
 \left\{
\begin{array}{ll}
- \beta\left(\left\Vert\emph{C}^{\prime}\right\Vert_{F} + \left\Vert\emph{P}^{\prime}\right\Vert_{F}\right)  & \emph{C}^{\prime}, \emph{P}^{\prime}\in\emph{V} \\
- \beta\left\Vert\emph{P}^{\prime}\right\Vert_{F} & \emph{C}^{\prime}\notin\emph{V} \\
- \beta\left\Vert\emph{C}^{\prime}\right\Vert_{F} & \emph{P}^{\prime}\notin\emph{V} \\
\end{array}
\right.$
\end{center}
where $\emph{V}$ is the set of optimizable variables of our model, and $\beta$ a hyper-parameter that in our experiments defaults to $\beta=0.3$.
We use a different regularizer in each alternative version of the model that we describe below.
These alternative versions have varying flexibility, i.e., either both $\emph{C}^{\prime}$ and $\emph{P}^{\prime}$ are optimizable variables ($\emph{C}^{\prime},\emph{P}^{\prime}\in\emph{V}$), or one of them is fixed, thus not optimizable ($\emph{C}^{\prime}\notin\emph{V}$ or $\emph{P}^{\prime}\notin\emph{V}$).
If both of them are fixed ($\emph{C}^{\prime},\emph{P}^{\prime}\notin\emph{V}$) then the model has no optimizable variables ($\emph{V} = \emptyset$).
Below we present the alternative versions of the model.


\subsubsection{Graph-Based Adaptation}
In this alternative (entitled \emph{GBA-CP}), we start with arbitrary cluster assignments for $\emph{C}^{\prime}$ and $\emph{P}^{\prime}$, which we both optimize based on the loss function.
This approach completely ignores the semantic information of $\emph{C}$ and $\emph{P}$ and adapts arbitrarily the clusters to the interconnection matrix $\emph{L}$.
This behavior of \emph{GBA-CP} is confirmed in our experiments (\S\ref{sec:results_clustering}).

In a less aggressive approach, we fix either $\emph{C}^{\prime}$ or $\emph{P}^{\prime}$ using one of the \emph{Content-Based} algorithms explained above, and optimize only one clustering (the non-fixed) based on the loss function.
We entitle these alternatives as \emph{GBA-C} for optimizing $\emph{C}^{\prime}$, and \emph{GBA-P} for optimizing $\emph{P}^{\prime}$ .

\subsubsection{Graph-Based Transformation}
In this alternative (entitled \emph{GBT-CP}), instead of optimizing directly $\emph{C}^{\prime}$ and $\emph{P}^{\prime}$, we optimize the weights of the non-linear neural transformations $\emph{f}_{\emph{C}}$ and $\emph{f}_{\emph{P}}$.
%
The architecture of $\emph{f}_{\emph{C}}$ and $\emph{f}_{\emph{P}}$ consists of a hidden layer of neurons with a rectified linear unit (\emph{ReLU}), and a linear \emph{Softmax} classifier that computes the overall cluster-membership distribution.
We use the same loss function as above where $\emph{C}^{\prime} = \emph{f}_{\emph{C}}(\emph{C})$ and $\emph{P}^{\prime} = \emph{f}_{\emph{P}}(\emph{P})$.
%
%

Similarly as above, in a less aggressive approach, we fix $\emph{C}^{\prime}$ or $\emph{P}^{\prime}$  using a \emph{Content-Based} algorithm, and optimize only the weights of one transformation ($\emph{f}_{\emph{C}}$ or $\emph{f}_{\emph{P}}$).
We entitle these alternatives as \emph{GBT-C} for optimizing $\emph{f}_{\emph{C}}$, and \emph{GBT-P} for optimizing $\emph{f}_{\emph{P}}$.


\subsection{Hybrid Clustering}\label{sec:coordinate_clustering}
The last clustering model that we propose is a \emph{Hybrid} model that combines a \emph{Content-Based} and a \emph{Graph-Based} model.
As we point out in our experimental evaluation (\S\ref{sec:results_clustering}), there is a trade-off between these two approaches in terms of the semantic and interconnection coherence of the computed clusters.
Hence, we introduce a tunable model that controls this trade-off.

Our model initializes the clusters $\emph{C}^{\prime}_{\emph{init}}$ and $\emph{P}^{\prime}_{\emph{init}}$ using a \emph{Content-Based} model.
Then, it uses an \emph{Alternate Optimization} (\emph{AO}) approach to jointly compute the final $\emph{C}^{\prime}$ and $\emph{P}^{\prime}$ that adjust best to $\emph{L}$.
More specifically, it iteratively freezes one of the two clusters and adjusts the other, until they both converge to an optimal state.
The loss function of this model is the following:
\begin{center}
$\emph{loss} = \left\{
\begin{array}{ll}
\gamma\left\Vert\emph{C}^{\prime} - \emph{L}\emph{P}^{\prime}\right\Vert_{F} + (1 - \gamma)\left\Vert\emph{C}^{\prime} - \emph{C}^{\prime}_{\emph{init}}\right\Vert_{F}  &  \emph{C}^{\prime}\text{-optim.}\\
\gamma\left\Vert\emph{C}^{\prime} - \emph{L}\emph{P}^{\prime}\right\Vert_{F} + (1 - \gamma)\left\Vert\emph{P}^{\prime} - \emph{P}^{\prime}_{\emph{init}}\right\Vert_{F}  & \emph{P}^{\prime}\text{-optim.}\\
\end{array}
\right.$
\end{center}
where $\gamma$ is a hyper-parameter that controls the trade-off between \emph{Content-Based} and \emph{Graph-Based} clustering.
In our experiments for brevity we present results for three values: \emph{AO-Content} for $\gamma=0.1$, \emph{AO-Balanced} for $\gamma=0.5$, and \emph{AO-Graph} for $\gamma=0.9$.


\section{Claim Contextualization}
\label{sec:verification}
%

In the previous section, we explain how we construct claim-paper clusterings in an unsupervised fashion.
These clusterings give already an initial context for claims since they relate them with relevant scientific literature.
In this section, we describe how we rank claims within clusters based on their check-worthiness and how we complement their fact-checking context by discovering (when available) previously verified related scientific claims.

%


\subsection{Check-Worthy Claim Ranking}
\label{subsec:in-cluster-ranking}

The check-worthiness of a scientific claim depends on its intent (e.g., whether it implies a causal relation or describes a particular aspect of an entity) and its prevalence (e.g., in news and social media).
We construct a custom in-cluster knowledge graph in which we encode the intent of the claims into the topology of the graph and the prevalence of the claims into the weighting of the graph.

%

%
%
%
%

\spara{In-Cluster Knowledge Graph.}
We construct a knowledge graph by using terms from a domain-specific vocabulary as nodes.
The edges of the graph denote the co-occurrence of two terms in the same claim
(e.g., the claim \emph{``Ibuprofen can worsen COVID-19 symptoms''} contributes the edge (\emph{Ibuprofen -- COVID-19})).

Since the dataset we use in our evaluation is health-related (details in \S\ref{sec:results}), we use the vocabulary of \emph{CDC A-Z Index}\footnote{\url{https://www.cdc.gov/az}} that includes health terms used by laypeople and professionals.
We note that the rest of the methodology is independent of the domain of the dataset, and can be simply adapted by selecting an appropriate vocabulary.

%

\spara{Graph Topology.}
We distinguish between two types of topologies based on two different intents:
\begin{itemize}[leftmargin=*]
	\item \emph{Causality-Based} topologies which contain nodes from distinct classes such as: 
	\begin{inparaenum}[i)]
		\item \emph{``Diseases and Disorders''} (e.g.,  \emph{Depression}, \emph{Influenza}, and \emph{Cancer}), and
		\item \emph{``Conditions, Symptoms, Medications, and Nutrients''} (e.g., \emph{Pregnancy}, \emph{Fever}, and \emph{Red Meat}). 
	\end{inparaenum}
	A directed edge between two nodes of a different class denotes, to a certain degree, a causal relation between these nodes \cite{DBLP:conf/chi/ChoudhuryKDCK16}.
	\item \emph{Aspect-Based} topologies which focus on the ``ego-network'' for one particular node (e.g., \emph{``COVID-19''}) and the different aspects regarding this node (e.g., \emph{``Origin''} or \emph{``Mortality Rate''}) \cite{DBLP:conf/aaai/MaPC18}.

\end{itemize}

%
%
%


%

%

\spara{Graph Weighting.}
The weighting scheme that we employ combines two criteria, namely the \emph{popularity} and the \emph{reputation} of the primary sources (i.e., the social media postings and the news articles) from which the claims were extracted.

The \emph{popularity} of a posting is computed as the sum of the number of re-postings and likes.
If multiple postings share the same claim, then their \emph{popularity} is aggregated.
Then, Box-Cox transformation ($\lambda=0$) \cite{10.2307/2984418}, to diminish the effect of the long-tail distribution, and Min-Max normalization in the interval $[0, 1]$ are applied.

On the other hand, the \emph{reputation} of a news article is entailed from the reputation of the news outlet that publishes the article.
In the context of this paper, we use the outlet scores compiled by the \emph{American Council on Science and Health} (\emph{ACSH}) ~\cite{acsh2017}, which we also normalize in the interval $[0, 1]$.
News outlets that are not on \emph{ACSH}'s list (i.e., ``long-tail'' outlets hosting only $13.5\%$ of the total articles in our collection) are assigned a neutral score ($0.5$).

Since we want to discover claims that are popular and come from low-reputable sources, we linearly combine the two metrics for each edge $e$, using a tuning parameter $\theta$ as follows:
\begin{center}
	$\emph{weight}(e) = \theta \; \emph{popularity}(e) + (1-\theta) \; (1-\emph{reputation}(e))$
\end{center}
In our implementation, we slightly favorite low reputation over popularity; thus, we use $\theta=0.4$.

\spara{Claim Ranking.}
We rank the edges, and consequently the claims, of the \emph{Causality-Based} topologies using the \emph{Betweenness Centrality} metric \cite{doi:10.1080/0022250X.2001.9990249}, and the \emph{Aspect-Based} topologies using the \emph{in-Degree} metric.
%
%
Examples of check-worthy claims in our data include the term pairs: (\emph{Autism}~--~\emph{Vaccines}), (\emph{Breast Cancer}~--~\emph{Abortion}), and (\emph{Chemotherapy}~--~\emph{Cannabis}) (details in \S\ref{sec:results_verification}).

\subsection{Enhanced Fact-Checking Context}\label{subsec:claims-context}
%
%
%
%


The final step for contextualizing the claims is to relate them (when available) with previously verified claims.
To retrieve such claims, we use \emph{ClaimsKG} \cite{DBLP:conf/semweb/TchechmedjievFB19}, a knowledge graph that aggregates claims and reviews published using \emph{ClaimReview}\footnote{\url{https://www.claimreviewproject.com}}.
%
After 
filtering out, based on the mentioned entities, claims with non-scientific content (i.e., $62.3\%$ of the total claims), we end up with a final set of $\textasciitilde4K$ scientific claims, out of which $79.8\%$ has been determined to be \emph{False}, and $20.2\%$ has been determined to be \emph{True}.
We relate claims by computing their Semantic Textual Similarity \cite{DBLP:conf/semeval/LiebeckPM016} and setting an appropriate threshold ($0.9$ in our experiments).
%

Our final fact-checking context for scientific claims consists of
related scientific papers and news articles from the same cluster, and, if available,
related, previously verified claims.
As we see in our experiments (\S\ref{sec:results_verification}), this enhanced context improves the verification accuracy and confidence of non-expert fact-checkers and helps them outperform commercial fact-checking systems.

\section{Experimental Evaluation}\label{sec:results}
In this section we evaluate the methods for
extraction (\S\ref{sec:results_extraction}),
clustering (\S\ref{sec:results_clustering}), and
contextualization (\S\ref{sec:results_verification})
of scientific claims.

\spara{Raw Dataset.}
We evaluate all three methods on a state-of-the-art dataset for measuring health-related scientific misinformation \cite{DBLP:conf/www/Smeros0A19}.
This dataset has the form of a directed graph, from \emph{social media postings} to \emph{news articles} to \emph{scientific papers}, where edges denote a hyperlink connection.
%
%
The $\textasciitilde50K$ \emph{social media postings} of the dataset include the text of the postings as well as popularity indicators such as the number of \emph{re-postings} and \emph{likes}.
The $\textasciitilde12K$ \emph{news articles} of the dataset include articles from mainstream news outlets (e.g., \rurl{theguardian.com} or \rurl{popsci.com}), as well as from alternative blogging platforms (e.g., \rurl{mercola.com} or \rurl{foodbabe.com}).
Finally, the $\textasciitilde24K$ \emph{scientific papers} of the dataset include peer-reviewed or gray literature\footnote{\url{https://en.wikipedia.org/wiki/Grey_literature}} papers hosted at universities, academic publishers, or scientific repositories (e.g., Scopus, PubMed, JSTOR, and CDC).
We note that the overall volume of the dataset simulates the typical news coverage on health-related topics for a period of four months.


\subsection{Evaluation of Claim Extraction}\label{sec:results_extraction}

The evaluation of the extractors is two-fold; first, we validate their accuracy using a widely-used clean and labeled dataset, and then, we use them in a real-world scenario where we apply them on the raw dataset described above, and evaluate them via crowdsourcing.
%

\subsubsection{Training}
\label{sec:training_data}

Since there is no specific training dataset for the task of scientific claim extraction, we use two datasets mainly used for argumentation mining, namely \emph{UKP} \cite{DBLP:conf/emnlp/StabMSRG18} and \emph{IBM} \cite{DBLP:conf/coling/LevyBHAS14}.
We train our classifiers using the balanced union of the two datasets ($\textasciitilde 11K$ positive and negative samples).
In the following, we refer to this dataset as the \emph{Generic Dataset} of claims.

We also train our classifiers with a ``science-flavored'' dataset derived from the \emph{UKP} and \emph{IBM} datasets.
Specifically, in this dataset, we oversample claims regarding, e.g., ``abortion''
and downsample claims regarding, e.g., ``school uniforms''.
%
We apply this data augmentation by manually processing based on the ``general topic'' field that exists in both \emph{UKP} and \emph{IBM} datasets.
The described dataset is also balanced, containing $\textasciitilde 16K$ positive and negative samples, and in the following, we refer to it as the \emph{Scientific Dataset} of claims.

%
%
%

\begin{table}[t]
	\centering
	\footnotesize
	\caption{Cross validation of scientific claim extractors. Since, as we explain in \S\ref{sec:training_data}, both datasets are balanced, the evaluation metric that we use is \emph{Accuracy} (\emph{ACC}).}
	\label{table:cross_validation}
	\begin{tabular}{clcc}
		\toprule
		{} & {} &  \textbf{Generic Dataset} & \textbf{Scientific Dataset} \\
		{} & {} & \emph{ACC} & \emph{ACC} \\
		\midrule
		\multirow{4}{*}{\tablevert{\textbf{Baseline}}}
		{} & \emph{Grammar-Based} &$50.4\%$  & $52.3\%$ \\
		{} & \emph{Context-Based} &  $49.5\%$ & $50.2\%$ \\
		{} & \emph{Random Forest} & $74.7\%$ & $75.6\%$  \\
		{} & \emph{BERT} & $\textbf{82.2\%}$ & $81.0\%$ \\
		\midrule
		\multirow{3}{*}{\tablevert{\textbf{SciClops}}}
		\smallskip
		{} & \emph{SciBERT} & $81.5\%$ & $80.6\%$ \\
		\smallskip
		{} & \emph{NewsBERT} & $82.0\%$ & $80.0\%$ \\
		\smallskip
		{} & \emph{SciNewsBERT} & $81.1\%$ & $\textbf{81.2\%}$ \\
		\bottomrule
	\end{tabular}
\end{table}

\subsubsection{Cross Validation}
We perform a 5-fold cross validation over the datasets described above; the results are shown in Table~\ref{table:cross_validation}.
We observe that the \emph{Heuristic-Based} extractors perform poorly for this task, which confirms that it is a demanding task with many corner cases.
%
Remarkably, the \emph{Context-Based} heuristic, which is domain-agnostic, achieves identical \emph{accuracy} with the \emph{Grammar-Based} heuristic, which contains manually curated grammar rules.
We also observe that the \emph{Random Forest} classifier does not perform extremely worse than the \emph{Transformer-Based} models, while being more eco-friendly in terms of resources and training time needed.

The performance of the transformer-based models confirms the fact that they are state-of-the-art in most NLP tasks.
However, from this task, we do not see the benefits of the domain-specific pretraining.
On the \emph{Generic Dataset}, \emph{BERT}, which is pre-trained on a generic corpus, performs better, while on the \emph{Scientific Dataset}, \emph{SciNewsBERT}, which is pre-trained on a scientific and a news corpus, performs better; nonetheless, their difference is negligible.
The real difference among these models is shown in the next experiment.

\subsubsection{Crowd Evaluation}

%
We collect boolean labels for $700$ sentences extracted from the raw dataset described above by asking the crowd workers a simple classification question (i.e., whether a given sentence contains a scientific claim or not).
We use the platform \emph{Mechanical Turk}, asking input from three independent crowd workers per sentence ($57$ in total).
To ensure high-quality annotations, we employ what the platform calls \emph{Master Workers}, i.e., the most experienced workers with approval rate greater than $80\%$.
%
%
Finally, we consider \emph{Strong Agreement} among crowd-workers, the 3 out of 3 agreement, and \emph{Weak Agreement} the 2 out of 3 agreement.

We note that there are $77$ out of the $700$ sentences for which the majority of the annotators answered \emph{N/A}, because they could not distinguish whether these sentences contain a claim or not.
For example, interrogative sentences like \emph{``What? Ibuprofen Can Make You Deaf?''} confused the annotators, while
similar affirmative sentences like \emph{``Tylenol PM Causes Brain Damage''} were easily identified as scientific claims.
%
The remaining $623$ sentences are divided into two subsets;
\begin{inparaenum}[i)]
	\item sentences having \emph{Strong Agreement} among annotators, with $82$ claims (positive examples) and $242$ non-claims (negative examples), and
	\item sentences having \emph{Weak Agreement} among annotators, with $125$ claims and $174$ non-claims.
\end{inparaenum}

We observe that especially the subset with \emph{Strong Agreement} is highly unbalanced, which is indeed a realistic scenario if we consider the ratio of claim and non-claim containing sentences in typical news articles.
Furthermore, annotators fully agree that a sentence contains a scientific claim for less than $12\%$ of total the sentences, which confirms it is a highly confusing task.

%
%

\begin{table}[t]
	\setlength{\tabcolsep}{3.5pt}
	\footnotesize
	\centering
	\caption{Crowd Evaluation of scientific claim extraction. Results reported for weak (2 out of 3) annotator agreement (\emph{125} claims - \emph{174} non-claims) and strong (3 out of 3) annotator agreement (\emph{82} claims - \emph{242} non-claims).
		Since, especially the second set is highly unbalanced, the evaluation metrics that we use are \emph{Precision} (\emph{P}), \emph{Recall} (\emph{R}), and \emph{F1 Score} (\emph{F1}).}
	\label{table:results_extraction}
	\begin{tabular}{clrrr|rrr}
		\toprule
		{} & {} & \multicolumn{3}{c}{\textbf{Weak Agreement}} & \multicolumn{3}{c}{\textbf{Strong Agreement}} \\
		{} & {} & \multicolumn{1}{c}{\emph{P}} & \multicolumn{1}{c}{\emph{R}}  & \multicolumn{1}{c}{\emph{F1}} & \multicolumn{1}{c}{\emph{P}} & \multicolumn{1}{c}{\emph{R}}  & \multicolumn{1}{c}{\emph{F1}} \\
		\midrule
		\multirow{6}{*}{\tablevert{\textbf{Baseline}}} 
		{} & \emph{Grammar-Based} & $51.8\%$ & $70.4\%$ & $59.9\%$ & $40.4\%$ & $28.0\%$ & $33.1\%$ \\
		{} & \emph{Context-Based} & $44.6\%$ & $49.6\%$ & $47.0\%$ & $24.5\%$ & $45.1\%$ & $31.8\%$ \\
		{} & \emph{Random Forest-gen} & $52.1\%$ & $70.4\%$ & $59.9\%$ & $43.7\%$ & $\textbf{80.5\%}$ & $56.7\%$ \\
		{} & \emph{Random Forest-sci} & $56.7\%$ & $54.4\%$ & $55.5\%$ & $43.3\%$ & $44.8\%$ & $44.1\%$ \\
		{} & \emph{BERT-gen} & $50.8\%$ & $50.4\%$ & $50.6\%$ & $33.5\%$ & $68.3\%$ & $45.0\%$ \\
		{} & \emph{BERT-sci} & $78.7\%$ & $38.4\%$ & $51.6\%$ & $79.2\%$ & $51.2\%$ & $62.2\%$ \\
		\midrule
		\multirow{6}{*}{\tablevert{\textbf{SciClops}}} 
		{} & \emph{NewsBERT-gen} & $55.0\%$ & $48.8\%$ & $51.7\%$ & $38.9\%$ & $62.2\%$ & $47.9\%$ \\
		{} & \emph{NewsBERT-sci} & $76.9\%$ & $40.0\%$ & $52.6\%$ & $74.2\%$ & $56.1\%$ & $63.9\%$ \\
		{} & \emph{SciBERT-gen} & $48.8\%$ & $66.4\%$ & $56.3\%$ & $32.8\%$ & $72.0\%$ & $45.0\%$ \\
		{} & \emph{SciBERT-sci} & $48.8\%$ & $66.4\%$ & $56.2\%$ & $\textbf{86.5\%}$ & $39.1\%$ & $53.8\%$ \\
		{} & \emph{SciNewsBERT-gen} & $49.8\%$ & $\textbf{80.0\%}$ & $\textbf{61.3\%}$ & $38.8\%$ & $78.0\%$ & $51.8\%$ \\
		{} & \emph{SciNewsBERT-sci} & $\textbf{84.4\%}$ & $30.4\%$ & $44.7\%$ & $82.7\%$ & $52.4\%$ & $\textbf{64.2\%}$ \\
		\bottomrule
	\end{tabular}
\end{table}

\spara{Results.}
The overall results of the comparison of the extraction models are summarized in Table~\ref{table:results_extraction}.
For all the models, we use the following naming convention: the suffix \emph{-gen} is used to denote that models are trained on the \emph{Generic Dataset} explained in \S\ref{sec:training_data}, while suffix \emph{-sci} is used to denote that models are trained on the \emph{Scientific Dataset} also explained in \S\ref{sec:training_data}.
This convention does not apply to heuristic models that do not require training.

We observe that all the \emph{-gen} models have better or equally good recall as the respective \emph{-sci} models.
This happens because \emph{-gen} models have been trained equally towards all the labeled claims and have learned to better recognize the structure of a claim.
After analyzing the errors of the models, we noticed that claims with simple structure like \emph{``Repetitive behaviors in autism show sex bias early in life''} were identified more from \emph{-gen} than from \emph{-sci} models.
On the other hand, \emph{-sci} models, which have been optimized for the narrow scientific domain, are more selective, hence they show in general better precision than the respective \emph{-gen} models.

Focusing more on the variants of \emph{BERT}, we observe that task-specific pretraining boosts the performance of the model, which is not visible in the first experiment.
Specifically, we see that pretraining on both scientific and news domain gives the best results.
One illuminative example is the claim \emph{``Galactosides Treat Urinary Tract Infections Without Antibiotics''}, where \emph{Galactosides} is a word that does not appear in the basic vocabulary of \emph{BERT}\footnote{\url{https://cdn.huggingface.co/bert-base-uncased-vocab.txt}}, however, it appears in the extended vocabulary of \emph{SciBERT}\footnote{\url{https://cdn.huggingface.co/allenai/scibert_scivocab_uncased/vocab.txt}} and \emph{SciNewsBERT}.

Finally, it is noteworthy that the \emph{Random Forest} model provides quite comparable results to the transformer-based models, while being, as stated above, a much lighter and faster-to-train model.

\subsection{Evaluation of Claim-Paper Clustering}\label{sec:results_clustering}
Since we construct a bimodal clustering of claims and papers, we evaluate its quality with respect to two axes;
a good-quality clustering must contain clusters of semantically related claims and papers (\emph{Semantic Coherence}), and adhere to the implicit connections between these claims and papers (\emph{Interconnection Coherence}).

\spara{Semantic Coherence.}
To measure the semantic coherence of a clustering, we compute a modified version of the \emph{Average Silhouette Width} (\emph{ASW}) \cite{ROUSSEEUW198753}.
The first modification is that the distance used is not a metric distance (e.g., Euclidean distance) but a semantic distance (Semantic Textual Similarity ($\emph{STS}$)).
%
%
The second modification is that we generalize the metric for two (or more) joint clusterings.
The original metric computes the average distance between the centroid of each cluster and its elements.
In our case, since we have two joint clusterings for claims and papers, we compute the metric for all the combinations of centroids ($\bar{c}$) and elements ($e$) of each cluster.
Thus, the modified \emph{ASW} is computed as follows:
%
\begin{center}
$\emph{ASW}(cluster) =
\frac{1}{|\operatorname{centroids}| \cdot |\operatorname{cluster}|} \smashoperator[r]{\sum_{\substack{e \in \operatorname{cluster} \\ \bar{c} \in \operatorname{centroids} }}} \emph{STS}(e, \bar{c})$
\end{center}
where \emph{centroids} consists of the claims centroid and the papers centroid of each cluster.
Finally, we report the mean $\emph{ASW}$ across all clusters.
%
%
This cross-computation of the metric allows capturing the semantic coherence of the clusters both individually and jointly.

\spara{Interconnection Coherence.}
To measure the interconnection coherence of the clusterings (i.e., the adaptivity of the clusterings towards the interconnection matrix $\emph{L}$), we use ideas from link-based recommendation.
First, we compute a hard clustering for claims and papers:
%
\begin{center}
$\begin{array}{ll}
	\emph{C}^{\prime}_{comp} &= argmax_x(\emph{C}^{\prime})\\
	\emph{P}^{\prime}_{comp} &= argmax_x(\emph{P}^{\prime})
\end{array}$

\end{center}
Since, as we explain in Table~\ref{table:clustering_notation}, each row of  $\emph{C}^{\prime}$ and $\emph{P}^{\prime}$ contains the probability of a claim or a paper to belong to a cluster, when we compute $argmax$ over rows we obtain a hard clustering, while when we compute $argmax$ over columns we obtain the cluster centroids.
For example, given a single claim $c$ and three clusters $cl_0, cl_1, cl_2$:
\begin{center}
$c^{\prime}= [0.1, 0.8, 0,1] \Rightarrow c^{\prime}_{comp} = cl_1$
\end{center}

Next, we use one clustering (e.g., of claims) to recommend possible instances of the other clustering (e.g., of papers).
The recommendation is content-agnostic and exploits only the interconnection matrix $\emph{L}$.
Formally:
\begin{center}
$\begin{array}{ll}
	\emph{C}^{\prime}_{rec} &= argsort_x(sum_y(\emph{L} \odot \emph{P}^{\prime}))\\
	\emph{P}^{\prime}_{rec} &= argsort_x(sum_y(\emph{L}^{\emph{T}} \odot \emph{C}^{\prime}))
\end{array}$
\end{center}
where $\odot$ is the Hadamard (element-wise) product.
%
For the same claim $c$, papers $p1$ and $p2$, and clusters $cl_0, cl_1, cl_2$ we have:
\begin{center}
$c
\begin{array}{ll}
\nearrow^{p_{1} [0.5,0.1,0.4]}\\
\searrow_{p_{2}[0.1,0.8,0.1]}\\
\end{array}
\Rightarrow
c^{\prime}_{rec} = argsort_x(0.6, 0.9, 0.5) = [cl_1, cl_0, cl_2]$
\end{center}


To compute the recommendation quality, we utilize the metric of \emph{Recall@k} (\emph{R@k}), which measures the ratio in which the correct cluster is recommended among the top-k results.
%
%
We report the mean of the \emph{R@k} for the claims and the papers clustering.

	\begin{table}[t]
	\setlength{\tabcolsep}{3.5pt}
	\footnotesize
	\centering
	\caption{
		Clustering Evaluation. \emph{Semantic Coherence} is measured using the \emph{Average Silhouette Width} (\emph{ASW}), and \emph{Interconnections Coherence} is measured using \emph{Recall@3} (\emph{R@3}).
	}
	\label{table:results_clustering}
	\begin{tabular}{cl|rr|rr|rr}
		\toprule
		\multicolumn{2}{c}{} & \multicolumn{2}{c}{\textbf{clusters=10}} &  \multicolumn{2}{c}{\textbf{clusters=50}} & \multicolumn{2}{c}{\textbf{clusters=100}}\\
		\multicolumn{2}{c}{} &  \multicolumn{1}{c}{ASW}  & \multicolumn{1}{c}{R@3} & \multicolumn{1}{c}{ASW} &  \multicolumn{1}{c}{R@3} & \multicolumn{1}{c}{ASW} & \multicolumn{1}{c}{R@3} \\
		\midrule
		\multirow{6}{*}{\tablevert{\textbf{Content-Based}}}
		{} & \emph{LDA}         &  44.5\% &   86.8\% &  63.2\% &   69.4\% &  66.6\% &  69.5\% \\
		{} & \emph{GSDMM}       &  42.1\% &   98.9\% & 48.5\% &   86.2\% &  48.7\% &  72.4\% \\
		{} & \emph{GMM}         &  55.5\% &   68.9\% & 67.7\% &   52.4\% &  72.8\% &  45.2\% \\
		{} & \emph{PCA/GMM}     &  51.3\% &   90.0\% & 66.6\% &   34.2\% &  71.7\% &  28.4\% \\
		{} & \emph{K-Means}     &  53.2\% &   97.9\% &  \textbf{68.9\%} &   83.4\% &  73.2\% &  74.2\% \\
		{} & \emph{PCA/K-Means} &  52.0\% &   97.6\% & 66.8\% &   87.8\% &  71.2\% &  75.1\% \\
		\midrule
		\multirow{6}{*}{\tablevert{\textbf{Graph-Based}}}
		{} & \emph{GBA-CP}      &  38.2\% &  \textbf{100.0\%} &  40.9\% &  \textbf{100.0\%} &  44.5\% &  99.5\% \\
		{} & \emph{GBA-C}       &  38.1\% &   96.7\% & 44.5\% &   93.2\% &  48.7\% &  92.0\% \\
		{} & \emph{GBA-P}       &  40.0\% &   96.5\% & 43.0\% &   93.6\% &  47.3\% &  92.3\% \\
		{} & \emph{GBT-CP}      &  26.5\% &   99.6\% & 27.1\% &   98.9\% &  32.1\% &  71.8\% \\
		{} & \emph{GBT-C}       &  37.9\% &   92.5\% & 45.0\% &   59.8\% &  47.2\% &  53.8\% \\
		{} & \emph{GBT-P}       &  36.4\% &   88.4\% & 42.3\% &   62.4\% &  43.7\% &  65.9\% \\
		\midrule
		\multirow{3}{*}{\tablevert{\textbf{Hybrid}}}
		{} & \emph{AO-Content}  &  54.8\% &   96.7\% & 67.9\% &   90.0\% &  \textbf{73.3\%} &  92.1\% \\
		{} & \emph{AO-Balanced} &  \textbf{56.0\%} &   99.8\% & 67.6\% &   99.6\% &  72.1\% &  99.5\% \\
		{} & \emph{AO-Graph}    &  55.6\% &   99.8\% & 67.3\% &  \textbf{100.0\%} &  71.8\% &  \textbf{99.8\%} \\
		\bottomrule
	\end{tabular}
\end{table}

\spara{Results.}
The results of the evaluation are shown in Table~\ref{table:results_clustering}.
As we observe, the \emph{Content-Based} (baseline) clustering techniques that use a textual representation of claims and papers (i.e., \emph{LDA} and \emph{GSDMM}), generate clusters with lower \emph{Semantic Coherence} than the ones that use an embeddings representation (i.e., \emph{GMM} and \emph{K-Means}).
This is partially explained by a vocabulary mismatch: the language used in papers is more complex and contains more scientific terms than the one used in social and news media (where the claims derive from).
Thus, embeddings representations have the advantage of capturing the semantic proximity of topics, even if these topics occur from two heterogeneous vocabularies.
Furthermore, we observe that soft clustering techniques (i.e., \emph{LDA} and \emph{GMM}) generate, in general, clusters with higher \emph{Semantic Coherence} than the respective hard clustering techniques (i.e., \emph{GSDMM} and \emph{K-Means}), indicating that the theme of claims and papers is usually multifaceted.
Finally, we observe that the dimensionality reduction, performed by \emph{PCA}, is not helpful in the context of this task.

Regarding the \emph{Graph-Based} techniques, we see that they construct clusters with high \emph{Interconnections Coherence} but the lowest \emph{Semantic Coherence}.
Not surprisingly, \emph{GBA-CP} achieves the maximum \emph{Interconnections Coherence} since, as we explain in \S\ref{sec:graph-based_clustering}, it arbitrarily adapts the clusters to the interconnection matrix $\emph{L}$.

Overall, we observe that the most robust technique in terms of balance between \emph{Semantic} and \emph{Interconnections Coherence} is the \emph{Hybrid} technique (\emph{AO-Balanced}), which computes a soft clustering based on an embeddings representation and considers both the text and the graph modality of the dataset equally.

\subsection{Evaluation of Claim Contextualization}
\label{sec:results_verification}

The overall evaluation of our method is performed with an experiment that involves expert and non-expert fact-checkers as well as two state-of-the-art commercial systems.
Using SciClops, we extract, cluster, and finally select the \emph{top-40} check-worthy scientific claims in the data collection.
The topics of the claims are heterogeneous, covering controversial online discussions such as the usage of therapeutic cannabis in modern medicine, the consumption of small amounts of alcohol during pregnancy, and the effect of vaccines in disorders such as autism.

\spara{Claim Post-Processing.}
We notice that in some of the claims, redundant information that could confuse the fact-checkers is mentioned (e.g., we find the claim ``Donald Trump has said vaccines cause autism,'' in which the scientific question is whether ``vaccines cause autism'' and not whether Donald Trump made this statement).
Thus, to avoid misinterpretations and to mitigate preexisting biases for or against public figures, we replace from these claims all the \emph{Person} and \emph{Organization} entities with indefinite pronouns.

\spara{Non-Experts.}
We employ crowdsourcing workers using the same setup described in \S\ref{sec:results_extraction}, and ask them to evaluate the \emph{Validity} of each claim in a \emph{Likert Scale} \cite{joshi2015likert} (from ``Highly Invalid'' to ``Highly Valid'').
We also ask them to rate their \emph{Effort} to find evidence and their \emph{Confidence} that the evidence they found is correct.

We divide non-experts into one control group of \emph{Non-Experts Without Context}, and two experimental groups of \emph{Non-Experts With Partial Context} and \emph{Non-Experts With Enhanced Context}:
\begin{itemize}[leftmargin=*]
	\item \emph{Non-Experts Without Context} are shown a bare scientific claim with no additional information, as they would read it online in, e.g., a messaging app or a social media posting.
	\item \emph{Non-Experts With Partial Context} are shown a scientific claim and its source news article, i.e., the news article from which the claim was extracted. 
	\item \emph{Non-Experts With Enhanced Context} are shown a scientific claim, its source news article, and:
	\begin{inparaenum}[i)]
		\item the top-k news articles where the same or similar claims were found,
		\item the top-k most relevant papers, and, if available,
		\item the top-k most similar, previously verified claims.
	\end{inparaenum}
	To avoid overwhelming this experimental group with redundant information, we set $k=3$.	
\end{itemize}

\spara{Experts.}
We ask two independent experts to evaluate the validity of the claims.
%
%
Each expert evaluated all $40$ claims independently, and was given the chance to cross-check the ratings by the other expert and revise their own ratings, if deemed appropriate.
Overall, we use the average of the two expert ratings as ground-truth. 

\spara{Commercial Systems.}
Finally, for the verification of the same scientific claims, we use two commercial systems for fact-checking, namely ClaimBuster \cite{DBLP:conf/kdd/HassanALT17} and Google Fact Check Explorer\footnote{\url{https://toolbox.google.com/factcheck/explorer}}:
\begin{itemize}[leftmargin=*]
	\item \emph{ClaimBuster} is a system used massively by journalists which initially aimed at detecting important factual claims in political discourses; however, its current architecture allows for investigating any kind of check-worthy claims (details in \S\ref{sec:relatedwork}).
	\item \emph{Google Fact Check Explorer} is also an exploration tool used by journalists to verify claims published using the tagging system of \emph{ClaimReview}; we note that \emph{ClaimReview} is also exploited in the contextualization step of SciClops (details in \S\ref{subsec:claims-context}).
\end{itemize}
To homogenize the scores of these systems with the scores of the fact-checkers, we quantize them to the aforementioned \emph{Likert Scale}.

\spara{Results.}
Results are summarized in Table~\ref{table:RMSE}. 
Given the ground-truth provided by the experts, we measure the accuracy of the three aforementioned groups of non-experts and the two commercial systems using the \emph{Root Mean Square Error} (\emph{RMSE}).

We observe that \emph{ClaimBuster} performs better than our control group of \emph{Non-Experts Without Context} while providing a solution without human intervention.
Furthermore, we observe that \emph{Google Fact Check Explorer} performs poorly, mainly because only $20\%$ of the queried claims were present in the fact-checking portals it monitors (e.g., the claim ``\emph{Vaccines cause Autism}'' is present in the fact-checking section of \emph{USA Today} \cite{usatoday2021}, while the \emph{Contradictory Claims} described next are absent from all the fact-checking portals).

Finally, regarding the non-expert human fact-checkers, we observe that the more contextual information is available, the more accurately they rate the claims.
Indicatively, the \emph{RMSE} of \emph{Non-Experts With Enhanced Context} is only $\emph{50\%}$ greater than the \emph{RMSE} across \emph{Experts}.
Overall, we see that, when the under-verification claims derive from a narrow scientific domain, \textbf{non-expert human fact-checkers, provided with the proper fact-checking context, may outperform state-of-the-art commercial systems}. 

%

%

\begin{table}[t]
	\footnotesize
	\centering
	\caption{Left: Root Mean Square Error (\emph{RMSE}) between the scores provided by the \emph{Experts} (ground-truth) and the scores provided by \emph{Non-Experts} and \emph{Commercial Systems}; the last row shows the \emph{RMSE} across \emph{Experts} (lower is better).
	\newline	 
	Right: Verification of two contradictory claims from \emph{CNN} and \emph{MensJournal} by \emph{Non-Experts} and \emph{Commercial Systems}; the last row shows the ground-truth provided by the \emph{Experts}.}
	\label{table:RMSE}
	\begin{tabular}{lc||cc}
		\toprule
		{} & \textbf{RMSE} & \textbf{CNN Claim} & \textbf{MensJournal Claim} \\
		\midrule		
		\textbf{Non-Experts} & {} & {} & {} \\
		\emph{Without Context} & $1.91$ & \emph{Borderline} & \emph{Borderline} \\
		\emph{With Partial Context}   & $1.73$ & \emph{\textbf{Valid}} & \emph{Valid} \\
		\emph{With Enhanced Context} & $\textbf{1.54}$ & \emph{\textbf{Valid}} & \emph{\textbf{Highly Invalid}} \\
		\midrule
		\textbf{Commercial Systems} & {} & {} & {} \\		
		\emph{ClaimBuster}   & $1.74$ & \emph{\textbf{Valid}} & \emph{Borderline} \\
		\emph{Google Fact Check Explorer}   & $2.79$ & \emph{N/A} & \emph{N/A} \\
		\midrule
		\textbf{Experts} & $1.02$ & \emph{\textbf{Highly Valid}} & \emph{\textbf{Highly Invalid}} \\
		\bottomrule
	\end{tabular}
\end{table}

\spara{Case Study: Contradictory Claims.}
%
%
%
%
Within the set of under-verification claims, we noticed two contradictory claims.
The first claim opposes the use of therapeutic cannabis for treating \emph{Post-Traumatic Stress Disorder} (\emph{PTSD}) and comes from a mainstream news outlet (\emph{CNN}).\footnote{\emph{CNN}: ``\emph{Marijuana does not treat chronic pain or post-traumatic stress disorder.}'' \cite{cnn2017}}
The second claim supports the use of cannabis for treating \emph{PTSD} and comes from a popular health blog (\emph{MensJournal}).\footnote{\emph{MensJournal}: ``\emph{Marijuana can help battle depression, anxiety, post-traumatic stress disorder, and even addictions to alcohol and painkillers.}'' \cite{mensjournal2017}}
Current scientific understanding supports the first claim (from \emph{CNN}), but not the second one (from \emph{MensJournal}), as evidenced by a paper of
the \emph{Journal of Clinical Psychiatry} \cite{Wilkinson2015}.

As we show in Table~\ref{table:RMSE}, \emph{ClaimBuster} and all the groups of \emph{Non-Experts} mostly support the claim from \emph{CNN} as valid. 
Moreover, as discussed above, \emph{Google Fact Check Explorer} provides no answer for these two claims since they are not present in the monitored fact-checking portals.
Indeed, only \emph{Non-Experts With Enhanced Context} were able to indicate that the claim from \emph{MensJournal} is invalid, mainly because \textbf{SciClops provided a fact-checking context that included the paper from the \emph{Journal of Clinical Psychiatry} which debunks the claim even in its title.\footnote{\emph{J. of Clinical Psychiatry}: ``\emph{Marijuana use is associated with worse outcomes in symptom severity and violent behavior in patients with posttraumatic stress disorder.}'' \cite{Wilkinson2015}}}
%

\begin{figure}[t]
	\centering
	\setlength{\tabcolsep}{3.5pt}
	\scriptsize
	\includegraphics[width=.57\linewidth]{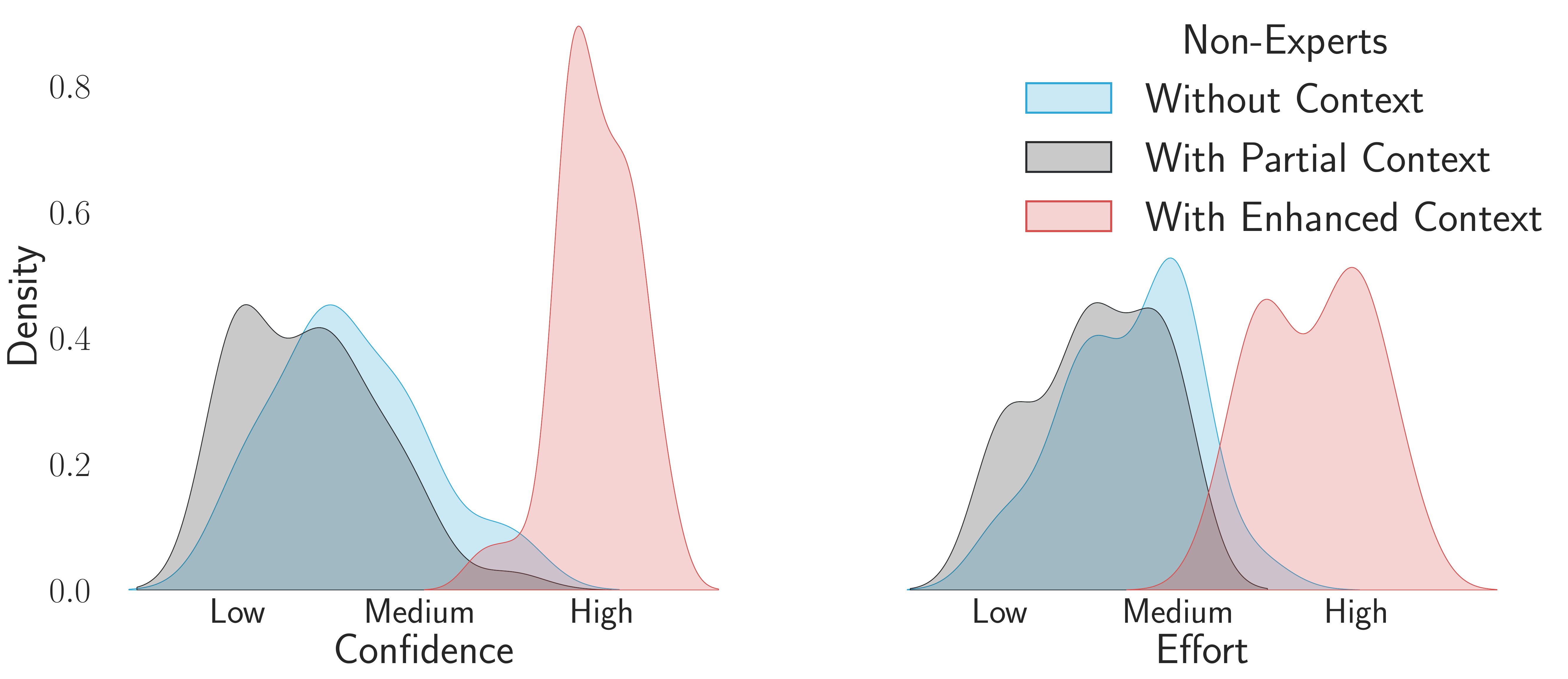}
	
	\begin{tabular}{ccc}
	\toprule
	\multicolumn{3}{c}{\textbf{Average Work Time of \emph{Non-Experts} (seconds)}}\\
	\emph{Without Context} & \emph{With Partial Context} & \emph{With Enhanced Context}\\
	$\emph{191s}$ & $\emph{388s}$ & $\emph{\textbf{898s}}$ \\
\end{tabular}
	\captionof{figure}{Kernel Density Estimation (\emph{KDE}) of \emph{Confidence} (left) and estimated \emph{Effort} (right), and \emph{Average Work Time} (bottom), of \emph{Non-Experts} verifying claims. Best seen in color.}
	\label{fig:KDE}
\end{figure}

\spara{Case Study: Confidence \& Effort.}
%
As we observe in Figure~\ref{fig:KDE}, \emph{Non-Experts} that were shown the \emph{Enhanced Context} of claims were more confident in their verification, additionally to being more accurate than the other two groups of users, which is partially explained by the fact that the provided context is fully-interpretable (as explained above), thus more trustworthy.
However, the same users' self-assessment of their effort as well as their actual work time was higher than the other two groups of users, which is explained by the fact that they had to visit more potential verification sources.
%
%

\section{Conclusions}\label{sec:conclusions}

We have presented an effective method for assisting non-experts in the verification of scientific claims.
We have shown that transformer models are indeed the state-of-the-art on scientific claim detection, however, they 
require domain-specific fine-tuning to perform better than other baselines.
%
We have also shown that, by exploiting the text of a claim and its connections to scientific papers, we effectively cluster topically-related claims and papers, as well as that, by building an in-cluster knowledge graph, we enable the detection of check-worthy claims.
Overall, we have shown that SciClops can build the appropriate fact-checking context to help non-expert fact-checkers verify complex scientific claims, outperforming commercial systems.
We believe that our method complements these systems in domains with sparse or non-existing ground-truth evidence, such as the critical domains of science and health.

\spara{Reproducibility.}
%
%
%
All the data, code, models, as well as expert and crowd annotations used for this paper are publicly available for research purposes in \textbf{\emph{\url{http://scilens.epfl.ch}}}.

\spara{Acknowledgments.}
We would like to thank the external experts: 
\emph{Dimitra Synefiaridou (Postdoctoral Researcher in Microbiology)} and
\emph{Sylvia Kyrilli (Senior Pediatrician)}.
This work has been partially supported by:
the \emph{Open Science Fund of EPFL};
the \emph{Swiss Academy of Engineering Sciences};
the \emph{HUMAINT project of the Joint Research Centre of European Commission};
the \emph{La Caixa project (LCF/PR/PR16/11110009)}.



\clearpage
\balance
\bibliographystyle{ACM-Reference-Format}
\bibliography{references}



\end{document}